\documentclass[12pt]{article}
\makeindex

\def\keywords#1{\small\centerline{\bf Keywords}\vspace{5mm}\centerline{\parbox{14cm}{#1}}}

\def\gtapprox{\buildrel{\lower.7ex\hbox{$>$}}\over
                       {\lower.7ex\hbox{$\sim$}}}
\def\nq{\hspace{-1em}}

\def\ignore#1{}

\def\qed{\sqcap\!\!\!\!\sqcup}
\def\odt{{\textstyle{1\over 2}}}

\def\hbar{h\!\!\!\!^{-}\,}

\def\beq{\begin{equation}}
\def\eeq{\end{equation}}
\def\beqn{\begin{displaymath}}
\def\eeqn{\end{displaymath}}
\def\bqa{\begin{equation}\begin{array}{c}}
\def\eqa{\end{array}\end{equation}}
\def\bqan{\begin{displaymath}\begin{array}{c}}
\def\eqan{\end{array}\end{displaymath}}
\def\pb{\underline}                       
\def\pb#1{\underline{#1}}                 

\begin{document}


\begin{titlepage}

\begin{center}  
 {\small Technical Report IDSIA-11-00, 14. November 2000}\\[2cm]

  {\Large\sc\hrule height1pt \vskip 2mm
  New Error Bounds for Solomonoff Prediction
\vskip 5mm \hrule height1pt  } \vspace{2cm}
  {\bf Marcus Hutter}                    \\[1cm]
  {\rm IDSIA, Galleria 2, CH-6928 Manno-Lugano, Switzerland}  \\
  {\rm\footnotesize marcus@idsia.ch \qquad http://www.idsia.ch$^{_{_\sim}}\!$marcus} \\[2cm]
\end{center}



\begin{abstract}
Solomonoff sequence prediction is a scheme to predict
digits of binary strings without knowing the underlying
probability distribution. We call a prediction scheme informed
when it knows the true probability distribution of the sequence.
Several new relations between universal Solomonoff sequence
prediction and informed prediction and general probabilistic
prediction schemes will be proved. Among others, they show that
the number of errors in Solomonoff prediction is finite for
computable distributions, if finite in the informed case.
Deterministic variants will also be studied. The most interesting
result is that the deterministic variant of Solomonoff prediction
is optimal compared to any other probabilistic or deterministic
prediction scheme apart from additive square root corrections
only. This makes it well suited even for difficult prediction
problems, where it does not suffice when the number of errors is
minimal to within some factor greater than one. Solomonoff's
original bound and the ones presented here complement each other
in a useful way.
\end{abstract}

\end{titlepage}

\section{Introduction}\label{int}
Induction is the process of predicting the future from the past
or, more precisely, it is the process of finding rules in (past)
data and using these rules to guess future data. The induction
principle has been subject to long philosophical controversies.
Highlights are Epicurus' principle of multiple explanations,
Occams' razor (simplicity) principle and Bayes' rule for
conditional probabilities \cite{Bay63}. In 1964, Solomonoff
\cite{Sol64} elegantly unified all these aspects into one formal
theory of inductive inference. The theory allows the prediction of digits
of binary sequences without knowing their true probability
distribution in contrast to what we call an informed scheme, where
the true distribution is known. A first error estimate was also
given by Solomonoff 14 years later in \cite{Sol78}. It states that
the total means squared distance of the prediction probabilities
of Solomonoff and informed prediction is bounded by the Kolmogorov
complexity of the true distribution. As a corollary, this theorem
ensures that Solomonoff prediction converges to informed
prediction for computable sequences in the limit. This is the key
result justifying the use of Solomonoff prediction for long
sequences of low complexity.

Another natural question is to ask for relations between the total
number of expected errors $E_\xi$ in Solomonoff prediction and the
total number of prediction errors $E_\mu$ in the informed scheme.
Unfortunately \cite{Sol78} does not bound $E_\xi$ in terms of
$E_\mu$ in a satisfactory way. For example it does not exclude the
possibility of an infinite $E_\xi$ even if $E_\mu$ is finite. Here
we want to prove upper bounds to $E_\xi$ in terms of $E_\mu$
ensuring as a corollary that the above case cannot happen. On the
other hand, our theorem does not say much about the convergence of
Solomonoff to informed prediction. So Solomonoff's and our bounds
complement each other in a nice way.

In the preliminary Section \ref{pre} we give some notations for
strings and conditional probability distributions on strings.
Furthermore, we introduce Kolmogorov complexity and the universal
probability, where we take care to make the latter a true
probability measure.

In Section \ref{ProbSP} we define the general probabilistic
prediction scheme ($\rho$) and Solo\-mo\-noff ($\xi$) and informed
($\mu$) prediction as special cases. We will give several error
relations between these prediction schemes. A bound for the error
difference $|E_\xi\!-\!E_\mu|$ between Solomonoff and informed
prediction is the central result. All other relations are then
simple, but interesting consequences or known results such as the
Euclidean bound.

In Section \ref{DetSP} we study deterministic variants of
Solomonoff ($\Theta_\xi$) and informed ($\Theta_\mu$) prediction.
We will give similar error relations as in the probabilistic case
between these prediction schemes. The most interesting consequence
is that the $\Theta_\xi$ system is optimal compared to any
other probabilistic or deterministic prediction scheme apart from
additive square root corrections only.

In the Appendices \ref{AppBasic2}, \ref{AppBasic1} and
\ref{AppBasic4} we prove the inequalities (\ref{basic2}),
(\ref{basic1}) and (\ref{basic4}), which are the central parts for
the proofs of the Theorems 1 and 2.

For an excellent introduction to Kolmogorov complexity and
Solomonoff induction one should consult the book of Li and
Vit\'anyi \cite{LiVi97} or the article \cite{LiVi92} for a short
course. Historical surveys of inductive reasoning/inference can be
found in \cite{Ang83,Sol97}.

\section{Preliminaries}\label{pre}

Throughout the paper we will consider binary sequences/strings and
conditional probability measures on strings.

We will denote strings over the binary alphabet $\{0,1\}$ by
$s\!=\!x_1x_2...x_n$ with $x_k\!\in\!\{0,1\}$ and their lengths
with $l(s)\!=\!n$. $\epsilon$ is the empty string,
$x_{n:m}:=x_nx_{n+1}...x_{m-1}x_m$ for $n\leq m$ and $\epsilon$
for $n>m$. Furthermore, $x_{<n}:=x_1... x_{n-1}$.

We use Greek letters for probability measures and underline
their arguments to indicate that they are probability arguments.
Let $\rho_n(\pb{x_1...x_n})$ be the probability that an (infinite)
sequence starts with $x_1...x_n$. We drop the index on
$\rho$ if it is clear from its arguments:
\beq\label{prop}
  \sum_{x_n\in\{0,1\}}\rho(\pb{x_{1:n}}) =
  \sum_{x_n}\rho_n(\pb{x_{1:n}}) =
  \rho_{n-1}(\pb{x_{<n}}) =
  \rho(\pb{x_{<n}})
  ,\quad
  \rho(\epsilon)=\rho_0(\epsilon)=1.
\eeq
We also need conditional probabilities derived from Bayes' rule.
We prefer a notation which preserves the order of the words in
contrast to the standard notation $\rho(\cdot|\cdot)$ which flips it. We extend the
definition of $\rho$ to the conditional case with
the following convention for its arguments: An underlined argument
$\pb{x_k}$ is a probability variable and other non-underlined
arguments $x_k$ represent conditions. With this convention, Bayes'
rule has the following look:
\bqa\label{bayes}
  \rho(x_{<n}\pb x_n) =
  \rho(\pb x_{1:n})/\rho(\pb x_{<n}) \quad\mbox{and}
  \\[4mm]
  \rho({\pb{x_1...x_n}}) =
  \rho(\pb x_1)\!\cdot\!
  \rho(x_1\pb x_2)
  \!\cdot\!...\!\cdot\!
  \rho(x_1...x_{n-1}\pb x_n).
\eqa
The first equation states that the probability that a string
$x_1...x_{n-1}$ is followed by $x_n$ is equal to the probability
that a string starts with $x_1...x_n$ divided by the probability
that a string starts with $x_1...x_{n-1}$. The second equation is
the first, applied $n$ times.

Let us choose some universal monotone Turing machine $U$ with
unidirectional input and output tapes and a bidirectional work
tape. We can then define the prefix Kolmogorov complexity \cite{Kol65,Lev74}
as the length of the shortest program $p$, for which $U$ outputs string $s$:
\beq\label{kolmodef}
  K(s) \;:=\; \min_p\{l(p): U(p)=s\}.
\eeq
The universal semi-measure $M(s)$ is defined as the probability
that the output of the universal Turing machine $U$ starts with
$s$ when provided with fair coin flips on the input tape. It is
easy to see that this is equivalent to the formal definition
\beq\label{uniM}
  M(s)\;:=\;\sum_{p\;:\;\exists\omega:U(p)=s\omega}\nq 2^{-l(p)},
\eeq
where the sum is over minimal programs $p$ for which $U$
outputs a string starting with $s$. $U$ might be non-terminating.
$M$ has the important universality property \cite{ZvLe70} that it
majorizes every computable probability measure $\rho$ up to a
multiplicative factor depending only on $\rho$ but not on $s$:
\beq\label{uni}
  \rho(\pb s)\leq 2^{K(\rho)+O(1)}M(s).
\eeq
The Kolmogorov complexity of a function like $\rho$ is defined as
the length of the shortest self-delimiting coding of a Turing
machine computing this function. Unfortunately $M$
itself is {\it not} a probability measure on the binary
strings. We have $M(s0)\!+\!M(s1)\!<\!M(s)$ because there are
programs $p$ which output just $s$, followed neither by $0$ nor
by $1$; they just stop after printing $s$ or continue forever
without any further output. This drawback can easily be
corrected\footnotemark \cite{Sol78}. Let us define the
universal probability measure $\xi$ by defining
first the conditional probabilities \beq\label{xidef}
  \xi(s\pb x) \;:=\; {M(sx)\over M(s0)+M(s1)}
  \quad,\quad
  x\in\{0,1\}
  \quad,\quad
  \xi(\epsilon):=1
\eeq
and then by using (\ref{bayes}) to get $\xi({\pb{x_1...x_n}})$.
It is easily verified by induction that $\xi$ is indeed a
probability measures and universal
\beq\label{unixi}
  \rho(\pb s)\leq 2^{K(\rho)+O(1)}\xi(\pb s).
\eeq
The latter follows from $\xi(\pb s)\geq M(s)$ and
(\ref{uni}). The universality property (\ref{unixi}) is all we
need to know about $\xi$ in the following.

\footnotetext{
Another popular way is to keep $M$ and sacrifice some of the
axioms of probability theory. The reason for doing this is that
$M$, although not computable \cite{LiVi97,Sol78}, is at least enumerable. On
the other hand, we are interested in conditional
probabilities, derived from $M$, which are no longer enumerable
anyway, so there is no reason for us to stick to $M$. $\xi$ is
still computable in the limit or approximable.
}

\section{Probabilistic Sequence Prediction}\label{ProbSP}

Every inductive inference problem can be brought into the
following form: Given a string $x$, give a guess for its
continuation $y$. We will assume that the strings which have to be
continued are drawn according to a probability
distribution\footnote{This probability measure $\mu$ might be $1$
for some sequence $x_{1:\infty}$ and $0$ for all others. In this case,
$K(\mu_n)$ is equal to $K(x_{1:n})$ (up to terms of order 1).}.
In this section we consider
probabilistic predictors of the next bit of a string. So let
$\mu(\pb{x_1... x_n})$ be the true probability measure of string
$x_{1:n}$, $x_k\!\in\!\{0,1\}$ and $\rho(x_{<n}\pb x_n)$ be the
probability that the system predicts $x_n$ as the successor of
$x_1...x_{n-1}$. We are not interested here in the probability of
the next bit itself. We want our system to output either $0$ or
$1$. Probabilistic strategies are useful in game theory where
they are called mixed strategies. We keep $\mu$ fixed and
compare different $\rho$. Interesting quantities are the
probability of making an error when predicting $x_n$, given
$x_{<n}$. If $x_n=0$, the probability of our system to predict $1$
(making an error) is $\rho(x_{<n}\pb 1)$$=$$1\!-\!\rho(x_{<n}\pb
0)$. That $x_n$ is $0$ happens with probability $\mu(x_{<n}\pb
0)$. Analogously for $0\!\leftrightarrow\!1$. So the probability of
making a wrong prediction in the $n^{th}$ step ($x_{<n}$ fixed) is
\beq\label{rhoerr}
  e_{n\rho}(x_{<n}) \;:=\; \sum_{x_n\!\in\{0,1\}}
  \mu(x_{<n}\pb x_n) [1-\rho(x_{<n}\pb x_n)].
\eeq
The total $\mu$-expected number of errors in the first $n$ predictions
is
\beq\label{toterr}
  E_{n\rho} \;:=\; \sum_{k=1}^n \nq\;\sum_{\quad x_1...x_{k-1}}\nq
  \mu(\pb x_{<k})\!\cdot\!e_{k\rho}(x_{<k}).
\eeq
If $\mu$ is known, a natural choice for
$\rho$ is $\rho=\mu$. This is what we call an informed prediction
scheme. If the probability of $x_n$ is high (low), the system
predicts $x_n$ with high (low) probability. If $\mu$ is unknown,
one could try the universal distribution $\xi$ for $\rho$ as
defined in (\ref{uniM}) and (\ref{xidef}). This is known as
Solomonoff prediction \cite{Sol64}.

What we are most interested in is an upper bound for the
$\mu$-expected number of errors $E_{n\xi}$ of the $\xi$-predictor.
One might also be interested in the probability difference of
predictions at step $n$ of the $\mu$- and $\xi$-predictor or the
total absolute difference to some power $\alpha$ ($\alpha$-norm in
$n$-space).
\beqn
  \displaystyle{
  d_k^\alpha(x_{<k}) \;:=\;
  \sum_{x_k}\mu(x_{<k}\pb x_k)\!\cdot\!
  \Big|\xi(x_{<k}\pb x_k)-\mu(x_{<k}\pb x_k)\Big|^\alpha \;=\;
  \Big|\xi(x_{<k}\pb 0)-\mu(x_{<k}\pb 0)\Big|^\alpha} \\[-4mm]
\eeqn
\beq\label{delta}
  \displaystyle{
  \Delta_n^{(\alpha)} \;:=\;
  \sum_{k=1}^n\sum_{\;\;x_{<k}}\mu(\pb x_{<k})\!\cdot\!
  d_k^\alpha(x_{<k}), \quad
  \alpha=1,2}
\eeq
For $\alpha\!=\!2$ there is the well known-result \cite{Sol78}
\beq\label{soleq}
\Delta_n^{(2)}\!<\odt\ln
2\!\cdot\!\!K(\mu)\!<\!\infty\quad \mbox{ for computable } \mu.
\eeq
One reason to directly study relations between $E_{n\xi}$ and $E_{n\mu}$ is
that from (\ref{soleq}) alone it does not follow that $E_{\infty\xi}$ is
finite, if $E_{\infty\mu}$ is finite.
Assume that we could choose $\mu$ such that
$e_{n\mu}\!\sim\!1/n^2$ and $e_{n\xi}\!\sim\!1/n$. Then
$E_{\infty\mu}$ would be finite, but
$E_{\infty\xi}$ would be infinite, without violating (\ref{soleq}).
There are other theorems, the most prominent
being $\xi(x_{<n}\pb x_n)/\mu(x_{<n}\pb
x_n)\stackrel{n\!\to\!\infty}\longrightarrow 1$ with $\mu$
probability 1 (see \cite{LiVi97} page 332). However, neither of
them settles the above question.
In the following we will show that a finite $E_{\infty\mu}$ causes a finite
$E_{\infty\xi}$.

Let us define the Kullback Leibler distance \cite{Kul59} or relative entropy
between $\mu$ and $\xi$:
\beq\label{hn}
  h_n(x_{<n}):=\sum_{x_n}\mu(x_{<n}\pb x_n)
  \ln{\mu(x_{<n}\pb x_n) \over \xi(x_{<n}\pb x_n)}.
\eeq
$H_n$ is then defined as the sum-expectation for which the
following can be shown \cite{Sol78}
\beq\label{entropy}
  H_n \;:=\; \sum_{k=1}^n \sum_{\;\;x_{<k}}
  \mu(\pb x_{<k})\!\cdot\!h_k(x_{<k}) \;=\;
  \sum_{k=1}^n\sum_{x_{1:k}}\mu(\pb x_{1:k})
  \ln{\mu(x_{<k}\pb x_k)\over\xi(x_{<k}\pb x_k)} \;=
\eeq
\\[-25pt]
\beqn
  =\;
  \sum_{x_{1:n}}\mu(\pb x_{1:n})
  \ln \prod_{k=1}^n{\mu(x_{<k}\pb x_k)\over\xi(x_{<k}\pb x_k)}
  \;=\;
  \sum_{x_{1:n}} \mu(\pb x_{1:n})
  \ln{\mu(\pb x_{1:n}) \over \xi(\pb x_{1:n})} \;\;
  <\;\;
  \ln 2\!\cdot\!K(\mu_n)+O(1)
\eeqn
In the first line we have inserted (\ref{hn}) and used Bayes rule
$\mu(\pb x_{<k})\!\cdot\!\mu(x_{<k}\pb x_k)\!=\!\mu(\pb x_{1:k})$.
Due to (\ref{prop}) we can replace $\sum_{x_{1:k}}\mu(\pb
x_{1:k})$ by $\sum_{x_{1:n}}\mu(\pb x_{1:n})$ as the argument of
the logarithm is independent of $x_{k+1:n}$. The $k$ sum can now
be exchanged with the $x_{1:n}$ sum and transforms to a product
inside the logarithm. In the last equality we have used the second
form of Bayes rule (\ref{bayes}) for $\mu$ and $\xi$. If we use
universality (\ref{unixi}) of $\xi$, i.e.\  $\ln \mu(\pb
x_{1:n})/\xi(\pb x_{1:n})\!<\!\ln 2\!\cdot\!K(\mu_n)\!+\!O(1)$, the
final inequality in (\ref{entropy}) is yielded, which is the basis of all
error estimates.

We now come to our first theorem:\\

{\sc Theorem 1.}
{\sl Let there be binary sequences $x_1x_2...$
drawn with probability $\mu_n(\pb x_{1:n})$ for the first $n$ bits. A
$\rho$-system predicts by definition $x_n$ from $x_{<n}$ with
probability $\rho(x_{<n}\pb x_n)$. $e_{n\rho}(x_{<n})$ is the
error probability in the $n^{th}$ prediction (\ref{rhoerr}) and
$E_{n\rho}$ is the $\mu$-expected total number of errors in the first $n$
predictions (\ref{toterr}). The following error relations hold
between universal Solomonoff ($\rho=\xi$), informed ($\rho=\mu$)
and general ($\rho$) predictions:
\beqn\label{th1}
\begin{array}{rl}
  i)   & |E_{n\xi}-E_{n\mu}| \quad\leq\quad
         \Delta_n^{(1)}      \quad<\quad
         H_n+\sqrt{2E_{n\mu}H_n} \\
  ii)  & \Delta_n^{(2)} \quad\leq\quad \odt H_n \\
  iii) & E_{n\xi} \quad>\quad \Delta_n^{(2)}+ \odt E_{n\mu} \\
  iv)  & E_{n\xi} \quad>\quad E_{n\mu}+H_n-\sqrt{2E_{n\mu}H_n}
         \quad>\; H_n \quad\quad\mbox{for}\quad E_{n\mu}
         \!>\!2H_n \\
   v)  & E_{n\mu} \quad\leq\quad 2E_{n\rho} \quad,\quad
         e_{n\mu} \;\leq\; 2e_{n\rho}
         \quad\;\mbox{for any }\rho \\
  vi)  & E_{n\xi} \quad<\quad 2E_{n\rho}+H_n+\sqrt{4E_{n\rho}H_n}
         \quad\mbox{for any }\rho,
\end{array}
\eeqn
where $H_n\!<\!\ln2\!\cdot\!K(\mu)\!+\!O(1)$ is the relative entropy
(\ref{entropy}) and $K(\mu)$ is the Kolmogorov complexity of $\mu$
(\ref{kolmodef}).}\\

{\sc Corollary 1.}
{\sl For computable $\mu$, i.e. for $K(\mu)<\infty$, the following
statements immediately follow from Theorem 1:
$$
\begin{array}{rl}
  vii)  & \mbox{if $E_{\infty\mu}$ is finite, then
         $E_{\infty\xi}$ is finite}
         \qquad\qquad\qquad\qquad\qquad\qquad\quad \\
 viii)  & E_{n\xi}/E_{n\mu} \quad\!=\; 1+O(E_{n\mu}^{-1/2})
          \stackrel{E_{n\mu}\to\infty}
         {\longrightarrow} 1 \\
   ix)  & E_{n\xi}-E_{n\mu} = \qquad O(\sqrt{E_{n\mu}}) \\
    x)  & E_{n\xi}/E_{n\rho} \quad\!\leq\; 2+O(E_{n\rho}^{-1/2}).
\end{array}
$$
}

Relation {\sl$(i)$} is the central new result. It is best illustrated for
computable $\mu$ by the corollary. Statements {\sl$(vii)$}, {\sl$(viii)$} and
{\sl$(ix)$} follow directly from {\sl$(i)$} and the finiteness of
$H_\infty$. Statement {\sl$(x)$} follows from {\sl$(vi)$}.

First of all, {\sl$(vii)$} ensures finiteness of the number of
errors of Solomonoff prediction, if the informed prediction makes
only a finite number of errors. This is especially the case for
deterministic $\mu$, as $E_{n\mu}\!=\!0$ in this case\footnote{We
call a probability measure deterministic if it is 1 for
exactly one sequence and 0 for all others.}. Solomonoff prediction
makes only a finite number of errors on computable sequences. For
more complicated probabilistic environments, where even the ideal
informed system makes an infinite number of errors, {\sl$(ix)$}
ensures that the error excess of Solomonoff prediction is only of
order $\sqrt{E_{n\mu}}$. This ensures that the error densities
$E_n/n$ of both systems converge to each other, but
{\sl$(ix)$} actually says more than this. It ensures that the quotient
converges to 1 and also gives the speed of convergence
{\sl$(viii)$}.

Relation {\sl$(ii)$} is the well-known Euclidean bound
\cite{Sol78}. It is the only upper bound in Theorem 1 which
remains finite for $E_{n\mu/\rho}\!\to\!\infty$. It ensures
convergence of the individual prediction probabilities
$\xi(x_{<n}\pb x_n)\!\to\!\mu(x_{<n}\pb x_n)$. Relation
{\sl$(iii)$} shows that the $\xi$ system makes at least half of
the errors of the $\mu$ system. Relation {\sl$(iv)$} improves the
lower bounds of {\sl$(i)$} and {\sl$(iii)$}. Together with the
upper bound in {\sl$(i)$} it says that the excess of $\xi$ errors
as compared to $\mu$ errors is given by $H_n$ apart from
$O(\sqrt{E_{n\mu}H_n})$ corrections. The excess is neither smaller
nor larger. This result is plausible, since knowing $\mu$ means
additional information, which saves making some of the errors. The
information content of $\mu$ (relative to $\xi$) is quantified in
terms of the relative entropy $H_n$.

Relation {\sl$(v)$} states that no prediction scheme can have less than
half of the errors of the $\mu$ system, whatever we take for
$\rho$. This ensures the optimality of $\mu$ apart from a factor
of 2. Combining this with {\sl$(i)$} ensures optimality of
Solomonoff prediction, apart from a factor of 2 and additive
(inverse) square root corrections {\sl$(vi)$}, {\sl$(x)$}. Note
that even when comparing $\xi$ with $\rho$, the computability of
$\mu$ is what counts, whereas $\rho$ might be any, even an
uncomputable, probabilistic predictor. The optimality within a
factor of 2 might be sufficient for some applications, especially
for finite $E_{\infty\mu}$ or if $E_{n\mu}/n\!\to\!0$, but is
inacceptable for others. More about this in the next section,
where we consider deterministic prediction, where no factor 2
occurs.

{\it Proof of Theorem 1.}
The first inequality in {\sl$(i)$} follows directly from the definition of $E_n$
and $\Delta_n$ and the triangle inequality. For the second
inequality, let us start more modestly and
try to find constants $A$ and $B$ which satisfy the linear inequality
\beq\label{Eineq2}
  \Delta_n^{(1)} \;<\; A\!\cdot\!E_{n\mu}+B\!\cdot\!H_n
\eeq
If we could show
\beq\label{eineq2}
  d_k(x_{<k}) \;<\; A\!\cdot\!e_{k\mu}(x_{<k}) +
                                B\!\cdot\!h_k(x_{<k})
\eeq
for all $k\leq n$ and all $x_{<k}$, (\ref{Eineq2}) would follow
immediately by summation and the definition of $\Delta_n$,
$E_n$ and $H_n$. With $k$, $x_{<k}$, $\mu$,
$\xi$ fixed now, we abbreviate
\beq\label{yzdef}
\begin{array}{lll}
  y:= \mu(x_{<k}\pb 1) &,& 1-y= \mu(x_{<k}\pb 0) \\
  z:= \xi(x_{<k}\pb 1) &,& 1-z= \xi(x_{<k}\pb 0) \\
  r:=\rho(x_{<k}\pb 1) &,& 1-r=\rho(x_{<k}\pb 0). \\
\end{array}
\eeq
The various error functions can then be expressed by $y$, $z$ and
$r$
\beq\label{abbr}
\begin{array}{rcl}
  e_{k\mu}  & = & 2y(1-y)       \\
  e_{k\xi}  & = & y(1-z)+(1-y)z \\
  e_{k\rho} & = & y(1-r)+(1-y)r \\
  d_k       & = & |y-z|         \\
  h_k       & = & y\ln{y\over z}+(1-y)\ln{1-y\over 1-z}. \\
\end{array}
\eeq
Inserting this into (\ref{eineq2}) we get
\beq\label{basic2}
  |y-z| \;<\; A\!\cdot\!2y(1-y)+
  B\!\cdot\!\left[y\ln{y\over z}+(1-y)\ln{1-y\over 1-z}\right].
\eeq
In Appendix \ref{AppBasic2}
we will show that this
inequality is true for $B\!\geq\!{1\over 2A}\!+\!1$, $A\!>\!0$.
Inequality (\ref{Eineq2}) therefore holds for any $A\!>\!0$,
provided we insert $B\!=\!{1\over 2A}\!+\!1$. Thus we might
minimize the r.h.s of (\ref{Eineq2}) w.r.t $A$. The minimum is at
$A=\sqrt{H_n/2E_{n\mu}}$ leading to the upper bound
$$
  \Delta_n^{(1)} \;<\; H_n+\sqrt{2E_{n\mu}H_n}
$$
which completes the proof of {\sl$(i)$}.

Bound {\sl$(ii)$} is well known \cite{Sol78}. It is already linear and
is proved by showing $d_n^2\!\leq\!\odt h_n$. Inserting the
abbreviations (\ref{abbr}) we get
\beq\label{kulbound}
 2(y-z)^2\;\leq\;y\ln{y\over z}+(1-y)\ln{1-y\over 1-z}
\eeq
This lower bound for the Kullback Leibler distance is well known
\cite{Kul59}.

Relation {\sl$(iii)$} does not involve $H_n$ at all and is elementary. It
is reduced to $e_{n\xi}\!>\!d_n^2\!+\!\odt e_{n\mu}$, equivalent
to $z(1-y)+y(1-z)>(y-z)^2+y(1-y)$, equivalent to $z(1-z)>0$, which
is obviously true.

The second inequality of {\sl$(iv)$} is trivial and the first is
proved similarly to {\sl$(i)$}. Again we start with a linear
inequality $ -E_{n\xi}\!<\!(A- 1)E_{n\mu}+(B- 1)H_n $, which is
further reduced to $-e_{k\xi}\!<\!(A-1)e_{k\mu}+(B-1)h_k$.
Inserting the abbreviations (\ref{abbr}) we get
\beq\label{basic1}
  -y(1-z)-z(1-y) \;<\; (A-1)2y(1-y)+
  (B-1)\left[y\ln{y\over z}+(1-y)\ln{1-y\over 1-z}\right].
\eeq
In Appendix \ref{AppBasic1} this
inequality is shown to hold for $2AB\!\geq\!1$, when $B\!>\!1$.
If we insert $B=1/2A$ and minimize w.r.t.\ $A$, the
minimum is again at $A=\sqrt{H_n/2E_{n\mu}}$ leading to the upper
bound
$
  -E_{n\xi} \!\leq\! -E_{n\mu}-H_n+\sqrt{2E_{n\mu}H_n}
$
restricted to $E_{n\mu}\!>\!2H_n$, which
completes the proof of {\sl$(iv)$}.

Statement {\sl$(v)$} is satisfied because $2y(1-y)\leq
2[y(1-r)+(1-y)r]$. Statement {\sl$(vi)$} is a direct consequence of
{\sl$(i)$} and {\sl$(v)$}. This completes the proof of Theorem 1.
$\qed$

\section{Deterministic Sequence Prediction}\label{DetSP}

In the last section several relations were derived between the
number of errors of the universal $\xi$-system, the informed
$\mu$-system and arbitrary $\rho$-systems. All of them were
probabilistic predictors in the sense that given $x_{<n}$ they
output $0$ or $1$ with certain probabilities. In this section, we
are interested in systems whose output on input $x_{<n}$ is
deterministically $0$ or $1$. Again we can distinguish between the
case where the true distribution $\mu$ is known or unknown. In the
probabilistic scheme we studied the $\mu$ and the $\xi$ system. Given
any probabilistic predictor $\rho$ it is easy to construct a
deterministic predictor $\Theta_\rho$ from it in the following way:
If the probability
of predicting $0$ is larger than $\odt$, the deterministic
predictor always chooses $0$. Analogously for $0\!\leftrightarrow\!1$.
We define\footnote{All results will be independent of the choice
for $\rho=\odt$, so one might choose $0$ for definiteness.}
$$
  \Theta_\rho(x_{<n}\pb x_n) \;:=\;
  \Theta(\rho(x_{<n}\pb x_n) - \odt)
  \;:=\; \left\{
  \begin{array}{c@{\quad\mbox{for}\quad}l}
    0                      & \rho(x_{<n}\pb x_n)<\odt \\
    1                      & \rho(x_{<n}\pb x_n)>\odt.
  \end{array} \right.
$$
Note that every deterministic predictor can be written in the form
$\Theta_\rho$ for some $\rho$ and that although
$\Theta_\rho(\pb{x_1...x_n})$, defined via Bayes' rule
(\ref{bayes}), takes only values in $\{0,1\}$, it may still be
interpreted as a probability measure. Deterministic
prediction is just a special case of probabilistic prediction. The
two models $\Theta_\mu$ and $\Theta_\xi$ will be studied now.

Analogously to the last section we draw binary strings randomly with distribution $\mu$
and define the probability that the $\Theta_\rho$ system makes an erroneous
prediction in the $n^{th}$ step and the total $\mu$-expected number of
errors in the first $n$ predictions as
\bqa\label{dtoterr}
  \displaystyle{
  e_{n\Theta_\rho}(x_{<n}) \;:=\; \sum_{x_n}
  \mu(x_{<n}\pb x_n) [1-\Theta_\rho(x_{<n}\pb x_n)] }
  \\[5mm]
  \displaystyle{
  E_{n\Theta_\rho} \;:=\; \sum_{k=1}^n \nq\;\sum_{\quad x_{<k}}
  \mu(\pb x_{<k})\!\cdot\!e_{k\Theta_\rho}(x_{<k}). }
\eqa
The definitions (\ref{hn}) and (\ref{entropy}) of $h_n$ and
$H_n$ remain unchanged ($\xi$ is {\sl not} replaced by
$\Theta_\xi$).

The following relations will be derived:\\

{\sc Theorem 2.}
{\sl Let there be binary sequences drawn with probability
$\mu_n(\pb x_{1:n})$ for the first $n$ bits. A $\rho$-system
predicts by definition $x_n$ from $x_{<n}$ with probability
$\rho(x_{<n}\pb x_n)$. A deterministic system $\Theta_\rho$ always
predicts $1$ if $\rho(x_{<n}\pb x_n)\!>\!\odt$ and 0 otherwise. If
$e_{n\rho}(x_{<n})$ is the error probability in the $n^{th}$
prediction, $E_{n\rho}$ the total $\mu$-expected number of errors
in the first $n$ predictions (\ref{toterr}), the following
relations hold:
\beqn\label{th2}
\begin{array}{rl}
  i)   & 0 \;\leq\; E_{n\Theta_\xi}-E_{n\Theta_\mu} \;=\;
         \sum_{x_k}\mu(\pb x_{<k})|e_{n\Theta_\xi}-e_{n\Theta_\mu}|
         \;<\; H_n+\sqrt{4E_{n\Theta_\mu}H_n+H_n^2} \\
  ii)  & E_{n\Theta_\mu}\leq E_{n\rho} \quad,\quad
         e_{n\Theta_\mu}\leq e_{n\rho} \qquad\qquad\qquad
         \mbox{for any }\rho \\
 iii)   & E_{n\Theta_\xi}
         \;<\; E_{n\rho}+H_n+\sqrt{4E_{n\rho}H_n+H_n^2} \qquad
         \mbox{for any }\rho, \\
\end{array}
\eeqn
where $H_n\!<\!\ln2\!\cdot\!K(\mu)\!+\!O(1)$ is the relative entropy
(\ref{entropy}), which is finite for computable $\mu$.
} 

No other useful bounds have been found, especially no bounds for the
analogue of $\Delta_n$.\\

{\sc Corollary 2.}
{\sl For computable $\mu$, i.e. for $K(\mu)<\infty$, the following
statements immediately follow from Theorem 2:
$$
\begin{array}{rl}
  vii)  & \mbox{if $E_{\infty\Theta_\mu}$ is finite, then
         $E_{\infty\Theta_\xi}$ is finite}
         \qquad\qquad\qquad\qquad\qquad\qquad\qquad
          \\
 viii)  & E_{n\Theta_\xi}/E_{n\Theta_\mu} \quad\!\!=\; 1+O(E_{n\Theta_\mu}^{-1/2})
          \longrightarrow 1 \quad\mbox{for}\quad E_{n\Theta_\mu}\to\infty \\
   ix)  & E_{n\Theta_\xi}-E_{n\Theta_\mu} =\qquad O(\sqrt{E_{n\Theta_\mu}}) \\
    x)  & E_{n\Theta_\xi}/E_{n\rho} \quad\;\leq\; 1+O(E_{n\rho}^{-1/2}).
\end{array}
$$
}

Most of what we said in the probabilistic case remains valid here,
as the Theorems and Corollaries 1 and 2 parallel each
other. For this reason we will only highlight the differences.

The last inequality of {\sl$(i)$} is the central new result in the
deterministic case. Again, it is illustrated in the corollary, which
follows trivially from Theorem 2.

From {\sl$(ii)$} we see that $\Theta_\mu$ is the
best prediction scheme possible, compared to any other
probabilistic or deterministic prediction $\rho$. The error
expectation $e_{n\Theta_\mu}$ is smaller in every single step and
hence, the total number of errors are also. This itself is not surprising
and nearly obvious, as the $\Theta_\mu$ system always predicts the bit
of highest probability. So, for known $\mu$, the $\Theta_\mu$ system
should always be preferred to any other prediction scheme, even to
the informed $\mu$ prediction system.

Combining {\sl$(i)$} and {\sl$(ii)$} leads to a bound
{\sl$(iii)$} on the number of prediction errors of the
deterministic variant of Solomonoff prediction. For computable
$\mu$, no prediction scheme can have fewer errors than that of the
$\Theta_\xi$ system, whatever we take for $\rho$, apart from some
additive correction of order $\sqrt{E_{n\Theta_\mu}}$. No factor 2
occurs as in the probabilistic case. Together with the quick
convergence $E_{n\rho}^{-1/2}$ stated in {\sl$(x)$}, the $\Theta_\xi$
model should be sufficiently good in many applications.

{\it Example.}
Let us consider a critical example. We want to predict the
outcome of a die colored black (=0) and white (=1). Two faces
should be white and the other 4 should be black.
The game becomes more interesting by having a second
complementary die with two black and four white sides. The dealer
who throws the dice
uses one or the other die according to some deterministic rule.
The stake $s$ is \$3 in every round; our return $r$ is \$5 for every
correct prediction.

The coloring of the dice and the selection strategy of
the dealer unambiguously determine $\mu$. $\mu(x_{<n}\pb 0)$ is
${2\over 3}$ for die 1 or ${1\over 3}$ for die 2.
If we use $\rho$ for prediction, we will have
made $E_{n\rho}$ incorrect and $n-E_{n\rho}$ correct predictions in the
first $n$ rounds. The expected profit will be
\beq\label{profit}
  P_{n\rho} \;:=\; (n-E_{n\rho})r-ns \;=\; (2n-5E_{n\rho})\$.
\eeq
The winning threshold $P_{n\rho}\!>\!0$ is reached if
$E_{n\rho}/n\!<\!1\!-\!s/r\!=\!{2\over 5}$.

If we knew $\mu$, we could use the best possible prediction scheme
$\Theta_\mu$. The error (\ref{dtoterr}) and profit (\ref{profit})
expectations per round in this case are
\beq\label{exethmu}
  e_{\Theta_\mu} \;:=\; e_{n\Theta_\mu}(x_{<n}) \;=\;
  {1\over 3} \;=\; {E_{n\Theta_\mu}\over n} \;<\; {2\over 5}
  \quad,\quad
  {P_{n\Theta_\mu}\over n} \;=\; {1\over 3}\$ > 0
\eeq
so we can make money from this game. If we predict according
to the probabilistic $\mu$ prediction scheme (\ref{rhoerr})
we would lose money in the long run:
\beqn
  e_{n\mu}(x_{<n}) \;=\; 2\!\cdot\!{1\over 3}\!\cdot\!{2\over 3}
  \;=\; {4\over 9} \;=\; {E_{n\mu}\over n} \;>\; {2\over 5}
  \quad,\quad
  {P_{n\mu}\over n} \;=\; -{2\over 9}\$ < 0
\eeqn
In the more interesting case where we do not know $\mu$ we can use
Solomonoff prediction $\xi$ or its deterministic variant
$\Theta_\xi$. From {\sl$(viii)$} of
Corollaries 1 and 2 we know that
\beqn
  P_{n\xi}/P_{n\mu} \;=\; 1+O(n^{-1/2}) \;=\;
  P_{n\Theta_\xi}/P_{n\Theta_\mu},
\eeqn
so asymptotically the $\xi$ system provides the same profit as the $\mu$ system and
the $\Theta_\xi$ system the same as the $\Theta_\mu$ system. Using the $\xi$
system is a losing strategy, while using the $\Theta_\xi$ system is a winning strategy.
Let us estimate the number of rounds we have to play before reaching the winning
zone with the $\Theta_\xi$ system. $P_{n\Theta_\xi}\!>\!0$ if
$E_{n\Theta_\xi}\!<\!(1\!-\!s/r)n$ if
\beqn
  E_{n\Theta_\mu} + H_n + \sqrt{4E_{n\Theta_\mu}H_n+H_n^2}
  \;<\; (1-s/r)\!\cdot\!n
\eeqn
by Theorem 2 {\sl$(i)$}. Solving w.r.t.\ $H_n$ we get
\beqn
  H_n \;<\; {(1-s/r-E_{n\Theta_\mu}/n)^2 \over
             2\!\cdot\!\!(1-s/r+E_{n\Theta_\mu}/n)} \cdot n.
\eeqn
Using $H_n\!<\!\ln 2\!\cdot\!K(\mu)\!+\!O(1)$ and
(\ref{exethmu}) we expect to be in the winning zone for
\beqn
  n \;>\; {2\!\cdot\!\!(1-s/r+e_{\Theta_\mu}) \over
  (1-s/r-e_{\Theta_\mu})^2} \cdot\!\ln 2\!\cdot\!K(\mu)+O(1)
  \;=\; 330\ln 2\!\cdot\!K(\mu)+O(1).
\eeqn
If the die selection strategy reflected in $\mu$ is not too
complicated, the $\Theta_\xi$ prediction system reaches the
winning zone after a few thousand rounds. The number of rounds is
not really small because the expected profit per round is one
order of magnitude smaller than the return. This leads to a
constant of two orders of magnitude size in front of $K(\mu)$.
Stated otherwise, it is due to the large stochastic noise, which
makes it difficult to extract the signal, i.e. the structure of
the rule $\mu$. Furthermore, this is only a bound for the turnaround
value of $n$. The true expected turnaround $n$ might be smaller.

However, every game for which there exists a winning strategy $\rho$ with
$P_{n\rho}\!\sim\!n$, $\Theta_\xi$ is guaranteed to get into
the winning zone for some $n\!\sim\!K(\mu)$, i.e.
$P_{n\Theta_\xi}\!>\!0$ for sufficiently large $n$. This is {\sl
not} guaranteed for the $\xi$-system, due to the factor $2$ in the
bound {\sl$(x)$} of Corollary 1.

{\it Proof of Theorem 2.}
The method of proof is the same as in the previous section, so
we will keep it short. With the abbreviations
(\ref{yzdef}) we can write $e_{k\Theta_\xi}$ and $e_{k\Theta_\mu}$
in the forms
\beq\label{dabbr}
\begin{array}{rcl}
  e_{k\Theta_\xi} & = & y(1-\Theta(z-\odt))+(1-y)\Theta(z-\odt)
                           \;=\; |y-\Theta(z-\odt)| \\
  e_{k\Theta_\mu} & = & y(1-\Theta(y-\odt))+(1-y)\Theta(y-\odt)
                           \;=\; \min\{y,1-y\}.          \\
\end{array}
\eeq
With these abbreviations,
{\sl$(ii)$} is equivalent to  $\min\{y,1-y\}\!\leq\!y(1-r)+(1-y)r$,
which is true, because the minimum of two numbers is always
smaller than their weighted average.

The first inequality and equality of {\sl$(i)$} follow directly
from {\sl$(ii)$}. To prove the last inequality, we start once
again with a linear model
\beq\label{eineq4}
  E_{n\Theta_\xi} \;<\; (A+1)E_{n\Theta_\mu} + (B+1)H_n.
\eeq
Inserting the definition of $E_n$ and $H_n$, using (\ref{dabbr}),
and omitting the sums we have to find $A$ and $B$, which satisfy
\beq\label{basic4}
  |y-\Theta(z-\odt)| \;<\; (A+1)\min\{y,1-y\}
  +(B+1)\left[y\ln{y\over z}+(1-y)\ln{1-y\over 1-z}\right].
\eeq
In Appendix \ref{AppBasic4}
we will show that the inequality is
satisfied for $B\!\geq\!{1\over 4}A+{1\over A}$ and $A>0$.
Inserting $B\!=\!{1\over 4}A+{1\over A}$ into (\ref{eineq4})
and minimizing the r.h.s. w.r.t.\ $A$, we get the upper bound
$$
E_{n\Theta_\xi} \;<\; E_{n\Theta_\mu} +
         H_n+\sqrt{4E_{n\mu}H_n+H_n^2}
\qquad\mbox{for}\qquad A^2={H_n\over E_{n\Theta_\mu}+ {1\over 4}H_n}.
$$
Statement {\sl$(iii)$} is a direct consequence of {\sl$(i)$} and
{\sl$(ii)$}. This completes the proof of Theorem 2. $\qed$

\section{Conclusions}
We have proved several new error bounds for Solomonoff prediction
in terms of informed prediction and in terms of general prediction
schemes. Theorem 1 and Corollary 1 summarize the results in the
probabilistic case and Theorem 2 and Corollary 2 for the
deterministic case. We have shown that in the probabilistic case
$E_{n\xi}$ is asymptotically bounded by twice the number of errors
of any other prediction scheme.
In the deterministic variant of Solomonoff prediction this factor 2 is absent.
It is well suited, even for difficult prediction problems, as the error
probability $E_{\Theta_\xi}/n$ converges rapidly to that of the
minimal possible error probability $E_{\Theta_\mu}/n$.

\paragraph{Acknowledgments:}
I thank Ray Solomonoff and J{\"u}rgen Schmidhuber for
proofreading this work and for numerous discussions.

\appendix
\section{Proof of Inequality (\ref{basic2})}\label{AppBasic2}

\footnote{The proofs are a bit sketchy. We will be a little sloppy
about boundary values $y=0/1$, $z=\odt$, $\Theta(0)$, $\geq$
versus $>$, and {\it approaching} versus {\it at} the boundary. All subtleties
have been checked and do not spoil the results. As $0\!<\!\xi\!<\!1$,
therefore $0\!<\!z\!<\!1$ is strict.}With the definition
\beqn
  f(y,z;A,B) \;:=\;
  A\!\cdot\!2y(1-y) +
  B\!\cdot\!\left[y\ln{y\over z}+(1-y)\ln{1-y\over 1-z}\right] -
  |y-z|
\eeqn
we have to show $f(y,z;A,B)\!>\!0$ for $0\!<\!y\!<\!1$,
$0\!<\!z\!<\!1$ and suitable $A$ and $B$. We do this by showing
that $f\!>\!0$ at all extremal values, `at' boundaries and at
non-analytical points. $f\!\to\!+\infty$ for $z\to 0/1$, if we
choose $B>0$. Moreover, at the non-analytic point $z=y$ we have
$f(y,y;A,B)=2Ay(1-y)\!\geq\!0$ for $A\!\geq\!0$. The extremal
condition $\partial f/\partial z\!=\!0$ for $z\!\neq\!y$
(keeping $y$ fixed) leads to
$$
  y \;=\; y^* \;:=\; z\!\cdot\![1-{s\over B}(1-z)],
  \quad s\;:=\;\mbox{sign}(z-y)\;=\;\pm 1.
$$
Inserting $y^*$ into the definition of $f$ and omitting the
positive term $B[\ldots]$, we get
\beqn
  f(y^*,z;A,B) \;>\; 2Ay^*(1-y^*)-|z-y^*| \;=\;
  \textstyle{{1\over B^2}z(1-z)}\!\cdot\!g(z;A,B)
\eeqn
\beqn
  g(z;A,B) \;:=\; 2A(B-s(1-z))(B+sz)-sB^.
\eeqn
We have reduced the problem to showing $g\!\geq\!0$. Since $s=\pm
1$, we have $g(z;A,B)>2A(B-1+z)(B-z)-B$ for $B>1$. The latter is
quadratic in $z$ and symmetric in $z\!\leftrightarrow\!1-z$ with a
maximum at $\odt$. Thus it is sufficient to check the boundary
values $g(0;A,B)=g(1;A,B)=2A(B-1)B-B$. They are non-negative for
$2A(B-1)\geq 1$. Putting everything together, we have proved that
$f\!>\!0$ for $B\!\geq\!{1\over 2A}+1$ and $A\!>\!0$. $\qed$

\section{Proof of Inequality (\ref{basic1})}\label{AppBasic1}

The proof of this inequality is similar to the previous one. With the definition
\beqn
  f(y,z;A,B) \;:=\;
  (A- 1)2y(1-y) +
  (B- 1)\left[y\ln{y\over z}+(1-y)\ln{1-y\over 1-z}\right] +
  y(1-z)+ z(1-y)
\eeqn
we have to show $f(y,z;A,B)\!>\!0$ for $0\!<\!y\!<\!1$,
$0\!<\!z\!<\!1$ and suitable $A$ and $B$. Again, we do this by showing
that $f\!>\!0$ at all extremal values and `at' the boundary.
$f\!\to\!+\infty$ for $z\to 0,1$, if we choose $B>1$. The
extremal condition $\partial f/\partial z\!=\!0$ (keeping $y$
fixed) leads to
$$
  y \;=\; y^* \;:=\; z\!\cdot\!{z- B\over 1- B-2z(1-z)},
  \quad 0<y^*<1.
$$
Inserting $y^*$ into the definition of $f$ and omitting the
positive term $(B- 1)[\ldots]$, we get
\beqn
  f(y^*,z;A,B) \;>\; 2Ay^*(1-y^*)- (2y^*-1)(z-y^*) \;=\;
  \textstyle{{z(1-z)\over[1- B-2z(1-z)]^2}}\!\cdot\!g(z;A,B)
\eeqn
\beqn
  g(z;A,B) \;:=\; 2A(z- B)(1-z- B)-(B- 1)(2z-1)^2.
\eeqn
We have reduced the problem to showing $g\!\geq\!0$. This is easy,
since $g$ is quadratic in $z$ and symmetric in
$z\!\leftrightarrow\!1-z$. The extremal value $g({1\over
2};A,B)$$=$$2A(B-\odt)^2$ is positive for $A\!>\!0$. The
boundary values $g(0;A,B)$$=$$g(1;A,B)$$=$$(2AB-1)(B- 1)$ are
$\geq 0$ for $2AB\geq 1$. Putting everything together, we have
proved that $f\!>\!0$ for $2AB\!\geq\! 1$ and $B\!>\!1$. $\qed$

\section{Proof of Inequality (\ref{basic4})}\label{AppBasic4}

We want to show that
$$
  \textstyle
  |y-\Theta(z-\odt)| \;<\;
  (A+1)\min\{y,1-y\} +
  (B+1)\left[y\ln{y\over z}+(1-y)\ln{1-y\over 1-z}\right]
$$
The formula is symmetric w.r.t. $y\!\leftrightarrow\!1\!-\!y$ and
$z\!\leftrightarrow\!1\!-\!z$ simultaneously, so we can restrict
ourselves to $0\!<\!y\!<\!1$ and $0\!<\!z\!<\!\odt$. Furthermore, let
$B\!>\!-1$. Using (\ref{kulbound}), it is enough to prove
$$
  f(y,z;A,B) \;:=\; (A+1)\min\{y,1-y\}+(B+1)2(y-z)^2-y \;>\; 0
$$
$f$ is quadratic in $z$; thus for $y\!<\!\odt$ it takes its
minimum at $z=y$. Since $f(y,y;A,B)=Ay>0$ for $A>0$, we can
concentrate on the case $y\geq\odt$. In this case, the
minimum is reached at the boundary $z=\odt$.
$$
  f(y,\odt;A,B) \;=\; (A+1)(1-y)+(B+1)2(y-\odt)^2-y
$$
This is now quadratic in $y$ with minimum at
$$
  y^* \;=\; {A+2B+4\over 4(B+1)}, \quad
  f(y^*,\odt;A,B) \;=\; {4AB-A^2-4\over 8(B+1)} \;\geq\; 0
$$
for $B\geq{1\over 4}A+{1\over A}$, $A>0$, $(\Rightarrow B\geq 1)$. $\qed$

\addcontentsline{toc}{section}{References}
\parskip=0ex plus 1ex minus 1ex

\end{document}